\documentclass{article}
\usepackage{ijcai21}

\usepackage{times}
\usepackage{soul}
\usepackage{url}
\usepackage[hidelinks]{hyperref}
\usepackage[utf8]{inputenc}
\usepackage[small]{caption}
\usepackage{graphicx}
\usepackage{amsmath}
\usepackage{amsthm}
\usepackage{booktabs}
\usepackage{algorithm}
\usepackage{algorithmic}

\usepackage{amsmath}

\DeclareMathOperator*{\argmin}{arg\,min}

\title{Substitutional Neural Image Compression}

\author{
Xiao Wang$^1$
\and
Wei Jiang$^2$\and
Wei Wang$^2$\And
Shan Liu$^2$\And
Brian Kulis$^1$\And
Peter Chin$^1$
\affiliations
$^1$Boston University\\
$^2$Tencent America\\
\emails
\{kxw, bkulis, spchin\}@bu.edu,
\{vwjiang, rickweiwang, shanl\}@tencent.com
}

\begin{document}

\maketitle

\begin{abstract}
  We describe Substitutional Neural Image Compression (SNIC), a general approach for enhancing any neural image compression model, that requires no data or additional tuning of the trained model. It boosts compression performance toward a flexible distortion metric and enables bit-rate control using a single model instance. The key idea is to replace the image to be compressed with a substitutional one that outperforms the original one in a desired way. Finding such a substitute is inherently difficult for conventional codecs, yet surprisingly favorable for neural compression models thanks to their fully differentiable structures. With gradients of a particular loss backpropogated to the input, a desired substitute can be efficiently crafted iteratively. We demonstrate the effectiveness of SNIC, when combined with various neural compression models and target metrics, in improving compression quality and performing bit-rate control measured by rate-distortion curves. Empirical results of control precision and generation speed are also discussed.
\end{abstract}

\section{Introduction}
Neural image compression (NIC) has recently emerged as a promising direction for advancing the state-of-the-art of lossy image compression \cite{toderici2015variable,balle2016end,shen2018codedvision,liu2019non,balle2018variational,toderici2017full,rippel2017real,mentzer2018conditional}. Besides outperforming modern engineered codecs such as JPEG~\cite{skodras2001jpeg}, JPEG 2000~\cite{rabbani2002jpeg2000} and BPG~\cite{bellard2015bpg} in rigorous distortion-based evaluation (e.g. PSNR), NIC has unique advantages in applications that are outside the conventional scope of image codecs, e.g. reconstructing realistic images from extremely low bit-rates by adding artificially synthesized image details \cite{agustsson2019generative,santurkar2018generative}.

\begin{figure*}[htbp]
  \centering
  \includegraphics[width=0.75\textwidth]{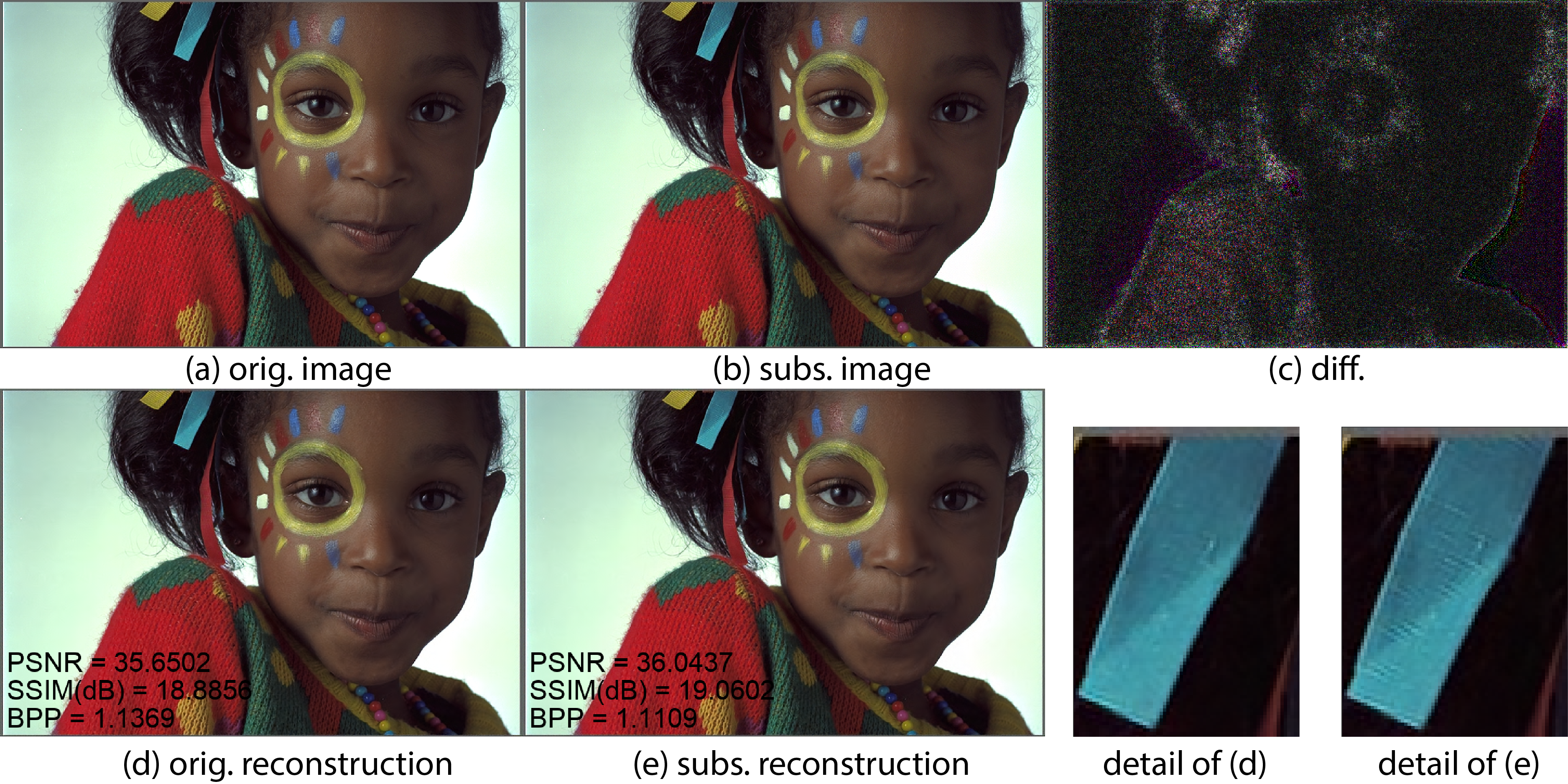}
  \caption{An example of compression results using the original image and the substitutioanl image. The difference image (c) has been scaled up for 100 times for visualization. In this example using substitutional image results in higher PSNR and MS-SSIM but lower BPP (bit per pixel). Note all distortions are measured against the original image. From the details of the blue band (bottom right), we see there are texture details preserved in (e), while in (d) texture details are completely lost. }
\label{fig:example}
\end{figure*}

NIC models feature a encoder-decoder style structure. An input image $x$ is first encoded into a compact latent representation $y$ by the encoder network (analysis transform), followed by a quantization step (we denote quantized $y$ as $\hat{y}$). The discrete $\hat{y}$ can be further compressed into a bit-stream with length $R(x)$ for efficient storage and transmission, using lossless entropy coding methods such as arithmetic coding. We can reconstruct a restored image $\hat{x}$ from $\hat{y}$ through a decoder network (synthesis transform). The reconstruction quality is usually measured by distortion metrics such as PSNR and MS-SSIM~\cite{wang2003multiscale} (we denote it as $D(\hat{x}, x)$).

Both encoder and decoder networks contain trainable parameters. However, applying end-to-end training directly is difficult due to the non-differentiable operations in quantization and entropy coding. To facilitate back-propagation, non-differentiable operations need to be replaced by differentiable ones in training. For instance, It is common to replace quantization with noise injection, and entropy coding with a parameterized entropy estimator. For clarity, we use $\tilde{x}$, $\tilde{y}$ and $R_e(x)$ to replace $\hat{x}$, $\hat{y}$ and $R(x)$ for the differentiable surrogate model in training.

Note that in lossy image compression, one needs to deal with two competing desires: better reconstruction quality measured by $D(\hat{x}, x)$ versus less bits consumption measured by $R_e(x)$. Therefore, a trade-off factor $\lambda$ is used in the training loss:

\begin{equation}
\label{eq:test}
    \lambda D(\tilde{x}, x) + R_e(x) .
\end{equation}

Training with a large $\lambda$ results in smaller distortion but more bits consumption, and vice versa. This is a key trade-off in lossy compression known as the rate-distortion (R-D) trade-off, and one important requirement for lossy compression is bit-rate control, i.e the capability of providing compressed images with varying quality and bit-rates based on practical needs. A desired compression algorithm should be able to perform bit-rate control smoothly, accurately and efficiently.

Bit-rate control, unfortunately, has been a challenging task for previous NIC methods. Conventionally, it requires multiple model instances trained from different $\lambda$ values both senders and receivers. Even so, bit-rate control is achieved at a very coarse level. On the one hand, it is very costly to train and store a large number of model instances for the sake of control precision (a single NIC model instance can take hundreds of MBs). On the other hand, it usually takes several rounds of trial-and-error before finding the right model instance to use, as the relation between $\lambda$ and the resulting bit-rate is unknown and varying per image.


Another limitation of NIC is the lack of adaptivity in the compression process compared with modern engineered codecs. For instance, BPG has a mechanism of dividing a block into sub-blocks by comparing the compression results of the root block and the sum of its children. Contrarily, NIC features a one-shot forward propagation and their parameters are fixed once trained. A better compression performance of NIC can be possibly achieved by adding feedback loops in the framework.


Motivated by addressing the above mentioned limitations, in this paper, we propose Substitutionary Neural Image Compression (SNIC). This approach is built on top of an arbitrary pre-trained NIC model, and enforced by the key idea of searching for a suitable substitutional image that replaces the original image to be compressed, which leads to a desired compression performance such as controlling the bit-rate and/or improving reconstruction quality (measured against the original image, not the substitutional image). A gradient-based method for finding the optimal substitute given a particular compression target is proposed. Since there exists a fully differentiable version of any NIC model for training, gradients with respect to the input image can be effectively obtained by backpropagation. This process does not need any change to the original NIC model or data. From a system perspective, it facilitates a feedback loop for the model's input. An example of substitutional image that achieves higher reconstruction quality with less bits consumption is illustrated in Fig. \ref{fig:example}.

One appealing property of SNIC is the flexibility of optimization target (loss), which does not have to be aligned with the loss used for training the underlying NIC model. Thus bit-rate control can be performed by generating suitable substitutes using the same pre-trained NIC model instance. Similarly, the choice of targeting distortion metric is also flexible for SNIC. In our experiments, we show SNIC is capable of improving MS-SSIM scores while the underlying models are originally optimized for mean squared error (MSE).

\section{Method}
\label{sec:method}

\subsection{The Optimal Substitutional Image}
\begin{figure}[ht]
  \centering
  \includegraphics[width=0.43\textwidth]{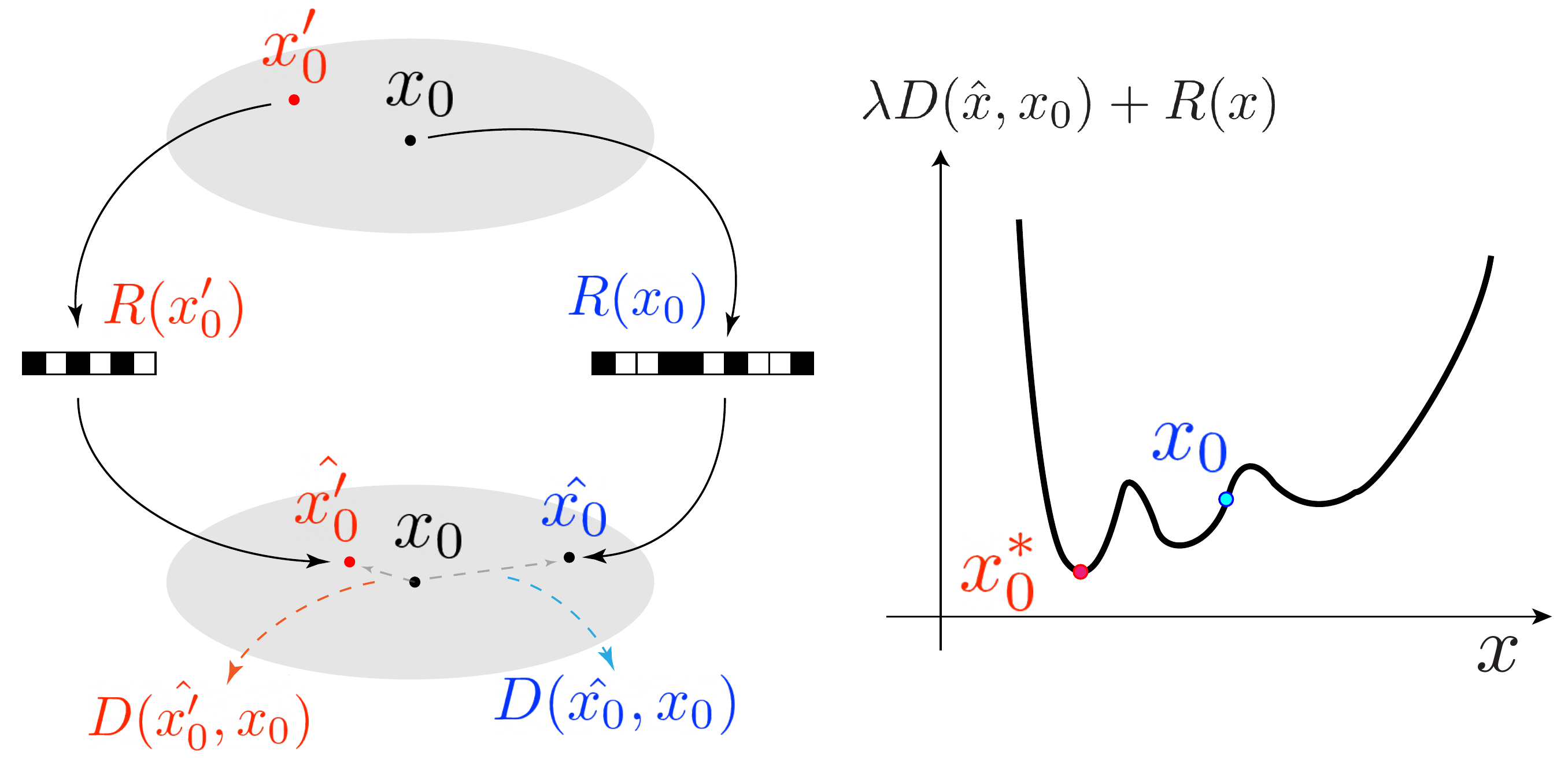}
  \caption{\textbf{Left}: compression and reconstruction by a specific compression model can be viewed as a 2-step mapping. If there exists a $x_0'$ such that it is mapped to $\hat{x_0'}$ that is closer to $x_0$ with $R(x_0') < R(x_0)$, then better compression can be achieved by simply substituting $x_0$ with $\hat{x_0'}$. \textbf{Right}: consider $\lambda D(\cdot, x_0) + R(\cdot)$ as a function of the input image. Best compression performance is achieved at the global minimum of this function.}
\label{fig:motivation}
\end{figure}

Neural image compression can be viewed as a two-step mapping process (see Fig. \ref{fig:motivation} left). 
The original image $x_0$ in a high dimensional space is mapped to a bit-stream with length $R(x_0)$ (the encoding mapping), which is then mapped back to the original space at $\hat{x_0}$ (the decoding mapping). We denote this 2-step mapping as a function $T(\cdot)$, i.e. $\hat{x_0}=T(x_0)$. The reconstruction quality is evaluated by a distortion metric $D(\hat{x_0}, x_0)$ .


The question is: When compressing a certain image $x_0$ , does applying the algorithm to $x_0$ necessarily yield the best compression result? For example, if there exists a substitute image $x_0'$, such that after applying the same two-step mapping, it is mapped to $\hat{x_0'}$ with $D(\hat{x_0'}, x_0) < D(\hat{x_0}, x_0)$, and at the same time $R(x_0')$ is smaller than $R(x)$, it means the compression performance for $x_0$ can be improved by replacing $x_0$ with $x_0'$ as the input of the compression model.

From another perspective, considering the minimization target $\lambda D(\hat{x}, x_0)+R(x)$ as a function of the input image $x$ (see Fig. \ref{fig:motivation} right), the original image is not necessarily at the global or even a local minimum of this function. 


Substitutional Neural Image Compression (SNIC) aims at finding the optimal substitute $x_0^*$ when compressing $x_0$, which can be formally expressed as

\begin{equation}
\label{eq:optimization_1}
    x_0^* := \argmin_x \lambda D(\hat{x}, x_0) + R(x) \ \  \text{s.t.} \ \  \hat{x}=T(x).
\end{equation}

It is worth noting that the reconstruction quality is still measured against the original image $x_0$, i.e. we seek to minimize $D(\hat{x},x_0)$ rather than minimizing $D(\hat{x}, x)$. 

For conventional codecs, the above problem is intrinsically difficult to solve, as searching in a high dimensional input space without heuristic is hopeless. However, for NIC, we already have a solvable alternative of the above optimization problem as follows.

\begin{equation}
\label{eq:optimization_2}
    x_0^* := \argmin_x \lambda_s D_s(\tilde{x}, x_0) + R_e(x) \ \  \text{s.t.} \ \  \tilde{x}=T_e(x)
\end{equation}

More specifically, as what we do in the NIC model training, we replace any non-differentiable operations with differentiable substitutes (we use $\tilde{x}=T_e(x)$ and $R_e(x)$ to indicate this change). As such, the optimal solution of problem \ref{eq:optimization_2} can be effectively found by iteratively updating $x_0$ via gradient descent.


Once $x_0^*$ is found, it does not require any change to neither the encoding network nor the decoding network. This approach is compatible with any pre-trained compression model, as long as it has a differentiable version, which is a precondition met by all neural compression models as it is also required for training.

\subsection{Bit-rate and Distortion Metric Control}
As mentioned above, the loss for generating substitutional images could be different from the loss used for training the underlying NIC model. In equation \ref{eq:optimization_2}, we use $\lambda_s$ to differentiate the trade-off factor in finding the substitute image from $\lambda$ in training. Bit-rate control is achieved by changing $\lambda_s$ and using different substitutes with different characteristics. Since $\lambda_s$ is a continuous variable, bit-rate can be adjusted smoothly. This is a significant advantage over conventional method through switching among different model instances. 

Nonetheless, it remains to investigate the reachable bit-rate range by using SNIC with a single model instance and what a R-D trade-off curve it generates. As SNIC also has the effect of improving compression performance, it can possibly enable flexible bit-rate control with a improved R-D curve over a certain bit-rate range. We show this is indeed the case with empirical results in the experiments section. 

Similarly, the target distortion metric in Eq. \ref{eq:optimization_2} can also be specified as desired, regardless of the target distortion metric used for training the underlying NIC model. Especially, we are interested in seeing how well SNIC improves compression toward a distortion metric that the underlying NIC model is not trained for. This part will also be further discussed in the experiments section.

\subsection{Direct Bit-rate and Distortion Control} 

Controlling R-D trade-off by specifying $\lambda_s$ can still be inconvenient, as the correspondence between $\lambda_s$ and the resulting bit-rate is unknown and possibly nonlinear. In many scenarios, this is a targeting bit-rate for compression decided by factors such as bandwidth. Therefore, it is desired to control the bit-rate of compressed images directly instead of searching for a suitable $\lambda_s$. For this purpose, SNIC can be used with a slightly modified formula:


\begin{equation}
\label{eq:optimization_3}
    \underset{x}{\text{minimize}} \quad D_s(T_e(x), x_0) + \kappa \max\{R_e(x) - R_t, \tau \}.
\end{equation}

In the above formula, $\kappa$ is a very large constant that puts a heavy penalty when $R_e(x)$ exceeds the target bit-rate $R_t$. Since gradient updates with a particular step-size could lead to an offset between real bit-rate $R_e(x)$ and target bit-rate $R_t$, we use a small value of $\tau$ to calibrate this offset. 

Similarly, distortion of compression can also be directly controlled if we switch $D$ and $R$ in the above formula, i.e.

\begin{equation}
\begin{aligned}
\label{eq:optimization_4}
    \underset{x}{\text{minimize}} \quad R_e(x) + \kappa \max\{D_s(T_e(x), x_0) - D_t, \tau \}.
\end{aligned}
\end{equation}

\subsection{Precision of Direct Bit-rate Control}
There are a number of factors that may influence the precision of direct bit-rate control using SNIC. First, the replacement of non-differentiable operations in substitutional image generation inevitably introduces errors. Second, as the optimization problem is solved via gradient descent, the optimal solution we found may not be the exact global minimum of the loss function. Third, for the flow of presentation, so far we assumed $x$ is a continuous variable. However in practice, the substitutional image will eventually be represented in the discrete space (e.g. 8-bit color RGB) with rounding errors. Therefore, we must demonstrate that the variance of the resulting bit-rate is sufficiently small in practice to justify our proposed method. Empirical results of this topic are also provided in the experiments section.




\subsection{Substitutional Latent Representation}
The optimization variable in formula \ref{eq:optimization_2}, \ref{eq:optimization_3} and \ref{eq:optimization_4} can also be any intermediate representations of the encoder before quantization and entropy coding. Optimizing over these intermediate representations can be faster, as it avoids propagating through the entire encoder. In this paper, we primarily study SNIC using the input image $x$ as the substitutional variable because (1) it provides insightful visualizations and (2) it provides baselines for comparison between different NIC models. However in practice, performance of SNIC may be further improved by choosing the best optimization variable given a specific model structure.

\section{Experiments}
\label{sec:experiment}

In this section, we demonstrate the effectiveness of SNIC in enhancing compression performance and bit-rate control in Sec.~\ref{sec:exp_enhance} and Sec.~\ref{sec:exp_bitratecontrol} respectively. It is important to note that the bit-rates in this section are real bit-rates using arithmetic coding instead of estimations. A study of the control precision of SNIC is presented in Sec.~\ref{sec:exp_controlprecision}. Computational cost of SNIC is presented in Sec.~\ref{sec:exp_speed}.


The testing dataset is Kodak\footnote{Downloaded from \texttt{http://r0k.us/graphics/kodak/}}, the mostly used benchmark dataset for image compression. We use two base NIC models from \cite{balle2016end} (indicated below as ICLR2017) and \cite{balle2018variational} (indicated below as ICLR2018), where we obtain four model instances for each of them via training with different $\lambda$ values ($0.001, 0.01, 0.1, 1$ for ICLR2017 and $0.001, 0.01, 0.1, 0.5$ for ICLR2018) in order to characterize their R-D trade-off curves as baselines. The training data is a subset of ImageNet~\cite{deng2009imagenet} and the training distortion metric is mean squared error (MSE). More training details are provided in appendix \ref{appendix:experiment_details}.

\subsection{Enhancing Compression Performance}
\label{sec:exp_enhance}

\begin{figure}[ht]
  \centering
  \includegraphics[width=0.48\textwidth]{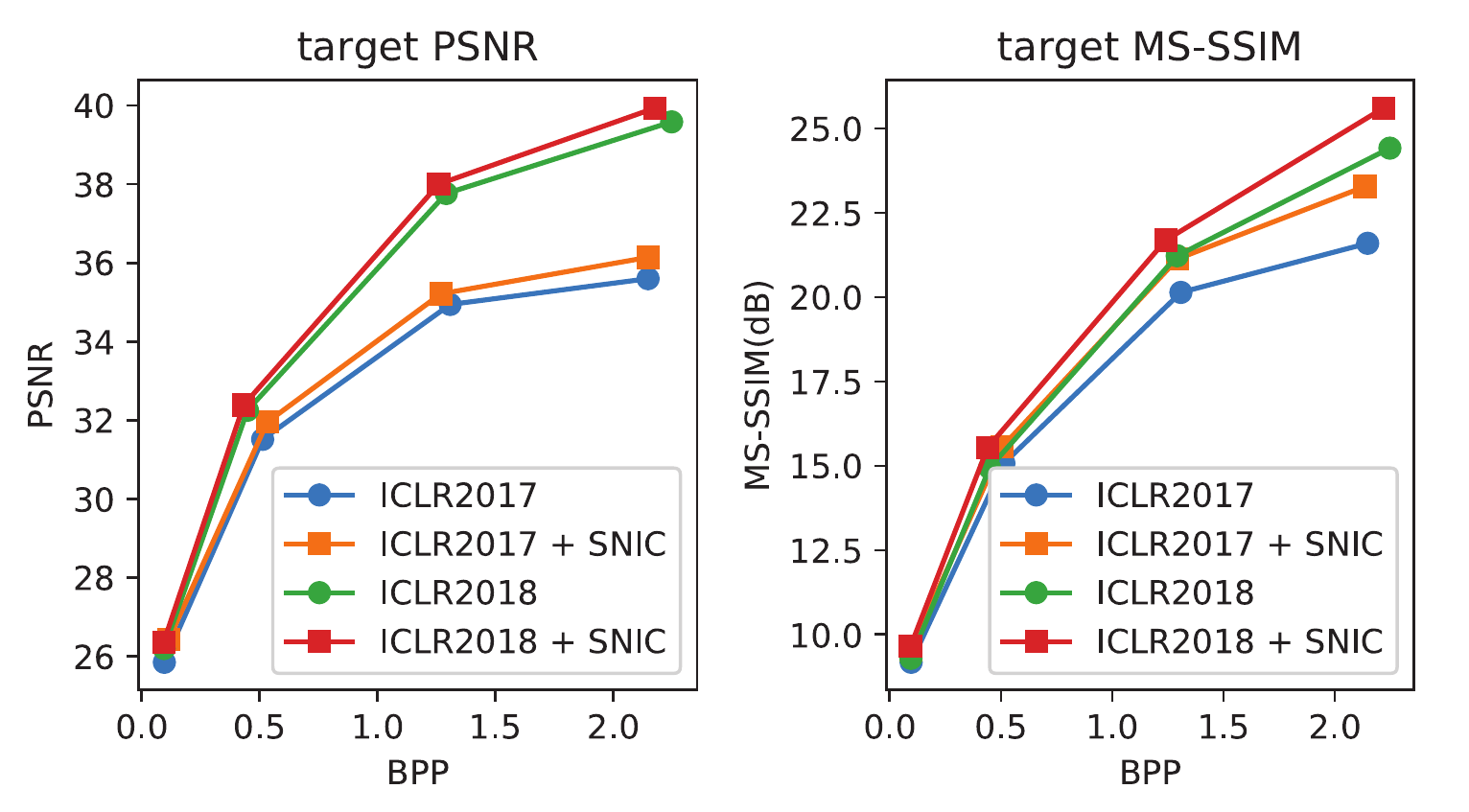}
  \caption{Comparison of R-D curves with and without SNIC (averaged over the Kodak dataset).}
\label{fig:e1_enhance}
\end{figure}

For both ICLR2017 and ICLR2018, we generate substitutional images for all trained model instances, so that a rigorous comparison between SNIC and baseline models can be measured by their R-D curves over a wide range of bit-rates. We conduct two sets of experiments that target improving PSNR and MS-SSIM respectively. When targeting PSNR, formula \ref{eq:optimization_2} is used to generate substitutes, where $\lambda_s$ is set to be the same as $\lambda$ of the underlying base model (i.e same loss used for both training and substitute generation). When targeting MS-SSIM, in order to align the bit-rates for a better comparison we use formula \ref{eq:optimization_3} where the target BPPs are the same as of original images.

All substitute images are generated from original images via a 100-step gradient descent process. Note that pixels are clipped after each gradient update step to ensure they are within the allowed perceptual range (i.e. $[0, 1]$). Additional experimental details are presented in appendix \ref{appendix:experiment_details}.

Fig.~\ref{fig:e1_enhance} shows compression performance measured by R-D trade-off curves using original images and substitutional images respectively. In all combinations of different target metrics and base models, SNIC leads to improved R-D curves. The improvement is more significant when it targets MS-SSIM where ICLR2017 with SNIC almost achieves the same R-D curve as ICLR2018 (note that ICLR2018 is 6 times larger than ICLR2017 in terms of parameters). Further more, we found that targeting PSNR and MS-SSIM have different effects on the other metric. Additional experimental results are presented in appendix \ref{appendix: non-targeting}.

In this experiment, we show that compression performance can be improved by solely changing the input of the compression model, with a flexible choice of target metric. Note that in the experiments we use the same setting (e.g. step size in gradient update) for all images and all model instances, but their values could also be treated as hyper-parameters to be optimized for a particular image-model pair. Unlike other machine learning tasks, hyper-parameter tuning at the inference phase is practical for SNIC, as the ground truth, which is the original image itself, is available. Although this leads to additional costs, it is still worth doing when the compressed images are used repeatedly that a small improvement leads to significant savings.

\subsection{Bit-rate Control}
\label{sec:exp_bitratecontrol}

\begin{figure}[ht]
  \centering
  \includegraphics[width=0.35\textwidth]{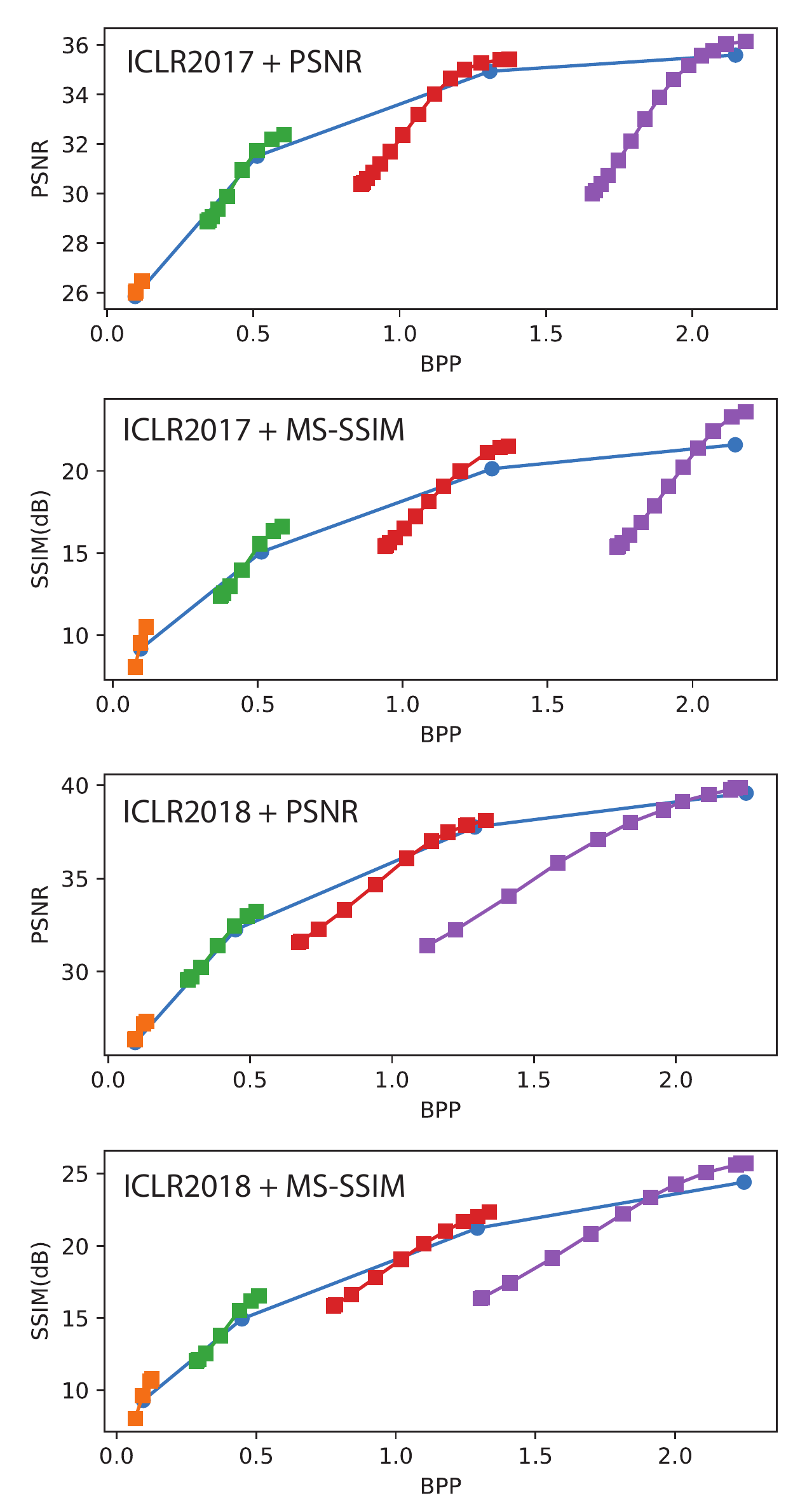}
  \caption{Achievable range of bit-rate control with different baseline models and target metrics. Points in the same color are generated with the same model instance using Eq.~\ref{eq:optimization_3}. The original R-D curve (blue line) is given by varying $\lambda$ in training which requires multiple models instances. The R-D curves are average over the Kodak dataset.}
\label{fig:exp_bitrate_control}
\end{figure}

Bit-rate control can be performed by either controlling trade-off factor $\lambda_s$ in Eq. \ref{eq:optimization_2} or by controlling target BPP $R_t$ directly in Eq. \ref{eq:optimization_3}. We found that the resulting R-D curves of using these two formula are similar, while Eq. \ref{eq:optimization_3} provides an easier way to have a equal spacing of points over the BPP range for demonstration purpose. The achievable range of bit-rate control and resulting R-D curves by using SNIC (Eq.~Eq. \ref{eq:optimization_3}) with a single model instance is shown in Fig. \ref{fig:exp_bitrate_control}. 

\begin{figure*}[t!]
  \centering
  \includegraphics[width=0.7\textwidth]{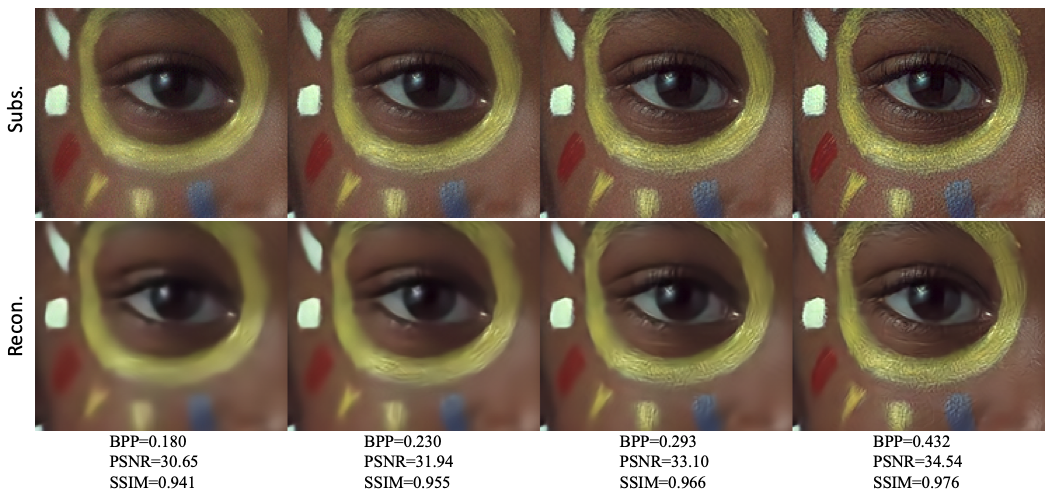}
  \caption{Substitutional images(clips) that target different bit-rates and their reconstructions. Substitutional images are generated from the same original image and the same compression model instance.}
\label{fig:exp_bitrate_control_example}
\end{figure*}

We observe that increasing and decreasing bit-rate using SNIC have different effects. When increasing bit-rate, both reconstruction quality and BPP increase, with a R-D curve above the baseline curve, before it reaches the upper limit, where a larger $R_t$ or $\lambda$ will not further increase neither BPP or reconstruction quality. The upper limit is only slightly above the original BPP. A larger controllable range is achieved for bit-rates below the original rate. However, as bit-rate decreases the R-D trade-off gets worse, which eventually makes the curve below the baseline curve. Also, the achievable control range using a single model instance is larger on high bit-rates than on low bit-rates.

Importantly, there is a non-trivial range of around $0.3$ where SNIC curve is above the baseline. In practice, we can use a few model instances with SNIC to cover the entire interested BPP range, while achieving superior compression performance. 



Another observation is that SNIC performs better with ICLR2018 compared to ICLR2017, not only with a wider control range, but also a better R-D trade-off. This indicates that although SNIC can be combined with any differentiable compression model, it is benefited from using more advanced ones.

An example of substitutional images and corresponding reconstructions targeting different bit-rates respectively is shown in Fig.~\ref{fig:exp_bitrate_control_example}. There is a visible difference between these images where the substitutes targeting for high bit-rates become more and more sharper. To some extent, SNIC enables image enhancement for better reconstruction quality and degradation for better compressibility, conditioned on the NIC model instance being used.

\subsection{Precision of Bit-rate Control}
\label{sec:exp_controlprecision}
In many applications, the goal of bit-rate control is to set the compression rate (BPP) to a particular target or range. Although BPP could be controlled by $\lambda_s$ in Eq.\ref{eq:optimization_2} (or conventionally by $\lambda$ in training multiple model instances), given that the correspondence between $\lambda_s$ and BPP is unknown, it may need several rounds of trial-and-error before reaching to the target range.

\begin{figure*}[ht]
  \centering\fontsize{9}{10}\selectfont
  \includegraphics[width=1\textwidth]{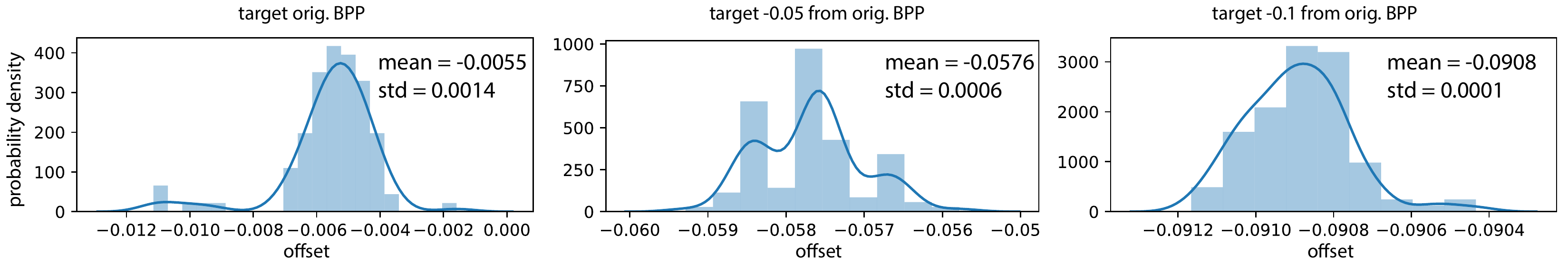}
  
  \begin{tabular}{l|l|l|l|l|l|l}
target BPP & orig. BPP & orig. - 0.05 & orig. - 0.1 & orig. -0.15 & orig. -0.2 & orig. - 0.25 \\ \hline
mean       & -0.0065   & -0.0676       & -0.1080 & -0.1303 & -0.1395 & -0.1419    \\ 
std        & 0.0014    & 0.0006       & 0.0005 & 0.0004 & 0.0002 & 0.0001      \\ 
\end{tabular}

\caption{Control precision measured by mean and standard deviation when $\tau=0$ in Eq.~\ref{eq:optimization_3}. \textbf{Above}: an example of BPP distributions with various BPP targets. Distributions are drawn by generating substitute images with the same target BPP for 100 times. The No.15 test image of Kodak dataset is used. \textbf{Below}: experimental results averaged on Kodak dataset.}
\label{fig:exp_control_precision}
\end{figure*}

Eq.~\ref{eq:optimization_3} provides a way to explicitly specify the target BPP. However, the real BPP may be different from the target due to several factors such as approximation, noise-injection and rounding errors. In this experiment, we seek to demonstrate the control precision of using Eq.~\ref{eq:optimization_3} measured by the distribution of resulting BPPs sampled repeated(see Fig. \ref{fig:exp_control_precision}). We particularly care about the variance of the distribution as it stems from uncontrollable factors such as noise and rounding errors. We demonstrate that the standard deviation is sufficiently small (mostly below $0.001$) in our experiments, indicating that the proposed method is effective. Note the mean could be calibrated by $\tau$ in Eq.~\ref{eq:optimization_3} at the cost of an additional round of generation (but still better that trial-and-error). 

\subsection{Speed of SNIC}
\label{sec:exp_speed}
As substitutional images are generated in an iterative manner, the speed of SNIC mainly depends on the number of gradient update steps and the base compression model being used. Theoretically, using more steps should be no worse than using less, so there exists a trade-off between performance and speed. In this experiment, we study the speed of substitute generation by comparing various numbers of steps and their effects on compression performance (see Tab. \ref{tab:generation_speed}). This experiment is conducted using \texttt{TensorFlow (V 1.15)} and a single \texttt{Nvidia GeForce RTX 2080Ti} graphics card.

\begin{table}[ht]
\caption{Average time consumption per image for generating substitutes and corresponding performance using different number of steps. Program time refers to the total time consumption including overheads such as variable initialization while generation time refers to only time for gradient updates. ICLR2017 is used as the base model.}
\label{tab:generation_speed}
\centering\fontsize{7}{8}\selectfont
\begin{tabular}{l|l|l|l|l|l|l}
\hline
Steps              & 0     & 5      & 10     & 50     & 100    & 500     \\ \hline
Program(s)       & -     & 0.159 & 0.223 & 0.778 & 1.488 & 7.129 \\ 
Generation(s)    & -     & 0.068 & 0.132 & 0.685 & 1.396 & 7.037 \\ \hline
\multicolumn{7}{l}{Performance(PSNR) on base models with different \textbf{training} $\lambda$}                                           \\ \hline
$\lambda$=0.001 & 25.84 & 25.90  & 25.96  & 26.23  & 26.32  & 26.42 \\  
$\lambda$=0.01  & 31.51 & 31.65  & 31.77  & 32.15  & 32.22  & 32.28   \\ 
$\lambda$=0.1   & 34.94 & 35.14  & 35.24  & 35.39  & 35.40  & 35.40   \\ 
$\lambda$=1     & 35.59 & 35.93  & 36.02  & 36.14  & 36.15  & 36.16   \\ \hline
\end{tabular}
\end{table}


It is observed that even a 5-step process provides a noticeable improvement, which takes less than $0.1$ second generation time. There is no much improvement after 100 steps. In practice, number of generation steps could be adjusted in real time easily based on practical needs as it requires no other change of the system.

\section{Related Works}

Inference time optimization of NIC has also been studied by Aytekin et al. \cite{aytekin2018block} and Campos et al. \cite{campos2019content}, where the authors finetune encoder weights and latent representations respectively on a per-image basis. Toward easier bit-rate control of NIC, Choi et al. and Yang et al. proposed to use conditional autoencoder and modulated autoencoder respectively \cite{choi2019variable,yang2020variable}. However, both approaches require special designs of the compression network and increase model complexity. In contrast, SNIC is a generic approach that can be combined with any differentiable model structure. Toward direct distortion control, Rozendaal et al. \cite{van2020lossy} add distortion constraints in the training phase. Contrarily, SNIC is able to control bit-rate or distortion directly in the inference phase. 

It is also worth mentioning that the technique of using input gradients appears in many other fields of deep learning, such as adversarial attack \cite{szegedy2013intriguing,carlini2017towards,madry2017towards} and feature visualization \cite{olah2017feature,mordvintsev2015inceptionism}. However, previous works focused mostly on classifiers, while in this paper we deal with an encoder-decoder network with two competing loss terms. 

\section{Discussion and Conclusion}
\label{sec:conclusion}

In this paper, we propose a general approach for accurate bit-rate control with boosted compression performance of NIC achieved by replacing the original image with subtly crafted substitutes. Crafting such substitutes could be formulated as solving an optimization problem via gradient descent, facilitated by the fully-differential nature of neural compression models. The most appealing property of SNIC is the flexibility in designing loss, where various substitutional images with desired characteristics can be generated from the same image. 

The fundamental reason that the proposed approach is feasible is that image compression is essentially a self-supervised process. Therefore, it can be assumed that evaluation feedback in the inference phase is available, which is fundamentally different from other machine learning tasks such as classification. 


We show that the optimal model input for compressing a particular image may not be the image itself, even for minimizing the original training loss. This is because the theoretical limit of compression performance is achieved when both the input image and model parameters are jointly optimized. However, any change to the model parameters (especially of the decoder) must also be transmitted to the receiver's end, leading to additional bits consumption. 


As discussed previously, the theoretical optimal substitute of an image is given by Eq.~\ref{eq:optimization_1}. We solve Eq.~\ref{eq:optimization_2} and \ref{eq:optimization_3} in practice because Eq.~\ref{eq:optimization_1} is not directly solvable. One key replacement is that we use noise injection to mimic the effect of quantization. Although it serves the purpose of making models differentiable, it also makes gradients noisy. Therefore, one important direction to improve SNIC is finding a better differentiable replacement of quantization. 

There are a lot of other possible extensions of SNIC. Although we study SNIC is the scope of image compression, the method can be extended naturally to other neural compression tasks for audio, video and data in general. Moreover, for certain applications, the strict distortion-based reconstruction quality metric can be relaxed. For instance, if the compressed images are known to be used for a certain task, we could replace the distortion loss with task-specific loss (e.g. accuracy for classification), making it a task-aware compression method.


\bibliographystyle{named}
\bibliography{ijcai21}

\clearpage
\appendix

\setcounter{figure}{0}
\setcounter{table}{0}
\makeatletter
\renewcommand{\thefigure}{A\arabic{figure}}
\renewcommand{\thetable}{A\arabic{table}}

\section{Implementation Details}
\label{appendix:experiment_details}

\subsection{Data Processing and Model Training}
\label{appendix:expdetails_train}
The training dataset contains images larger than $256 \times 256$ pixels from ImageNet, which has around 20000 images in total. In the training data pipeline, each image is randomly cropped to a $256 \times 256$ patch. This step is important to prevent over-fitting. The training process is performed for 1000000 steps with a batch size of 8. The same training process is applied to both ICLR2017 model and ICLR2018 model.

\subsection{Substitutional Image Generation}
\label{appendix:expdetails_generate}
All substitutional images are generated from iteratively updating the input image using input gradients. This process starts with initializing the input image as the original image to be compressed. Gradient descent is performed for 100 steps, unless specified otherwise. The step size is $0.001$ when targeting PSNR and $0.0005$ when targeting MS-SSIM, as we found these settings work well in general. One exception exists in Tab.~\ref{tab:generation_speed} we use the best step size chosen from [0.0001 0.001 0.01 0.1 1]. There is a clipping step after each gradient update step in order to assure the updated image is within the valid perceptual range (i.e. [0, 1]). 

\section{Effects to the Non-targeting Metric}
\label{appendix: non-targeting}

\begin{figure}[ht]
    \centering
    \includegraphics[width=0.45\textwidth]{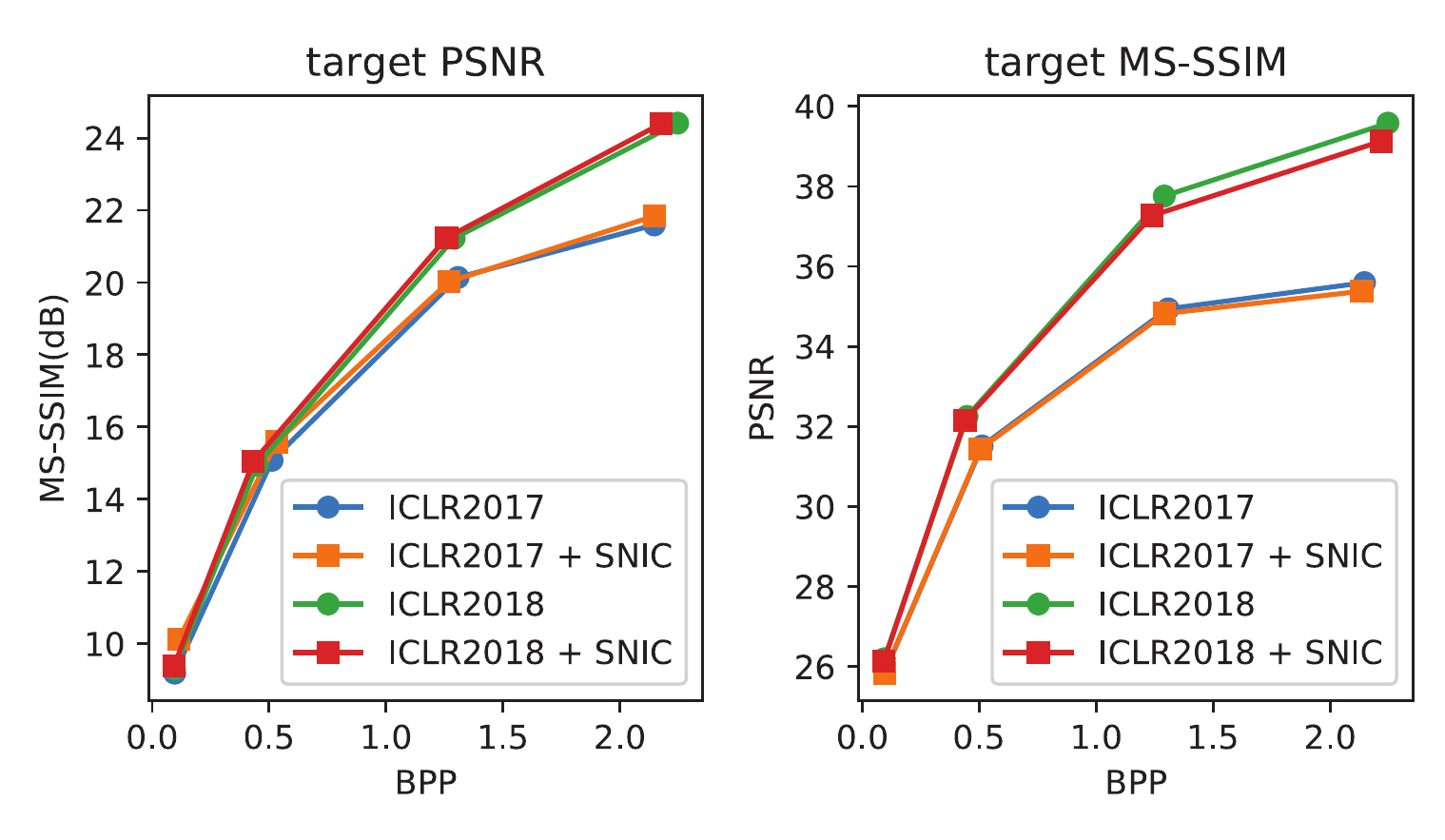}
    \caption{Effects of SNIC on the non-target distortion metric.}
    \label{fig: appendix_different_metric}
\end{figure}

When targeting PSNR, SNIC does not only improves PSNR but also improves MS-SSIM slightly. When targeting MS-SSIM, the improvement on MS-SSIM is more significant but it also leads to a small drop of PSNR on the high bit-rates (see Fig.~\ref{fig: appendix_different_metric}).

\begin{figure}[ht]
    \centering
    \includegraphics[width=0.45\textwidth]{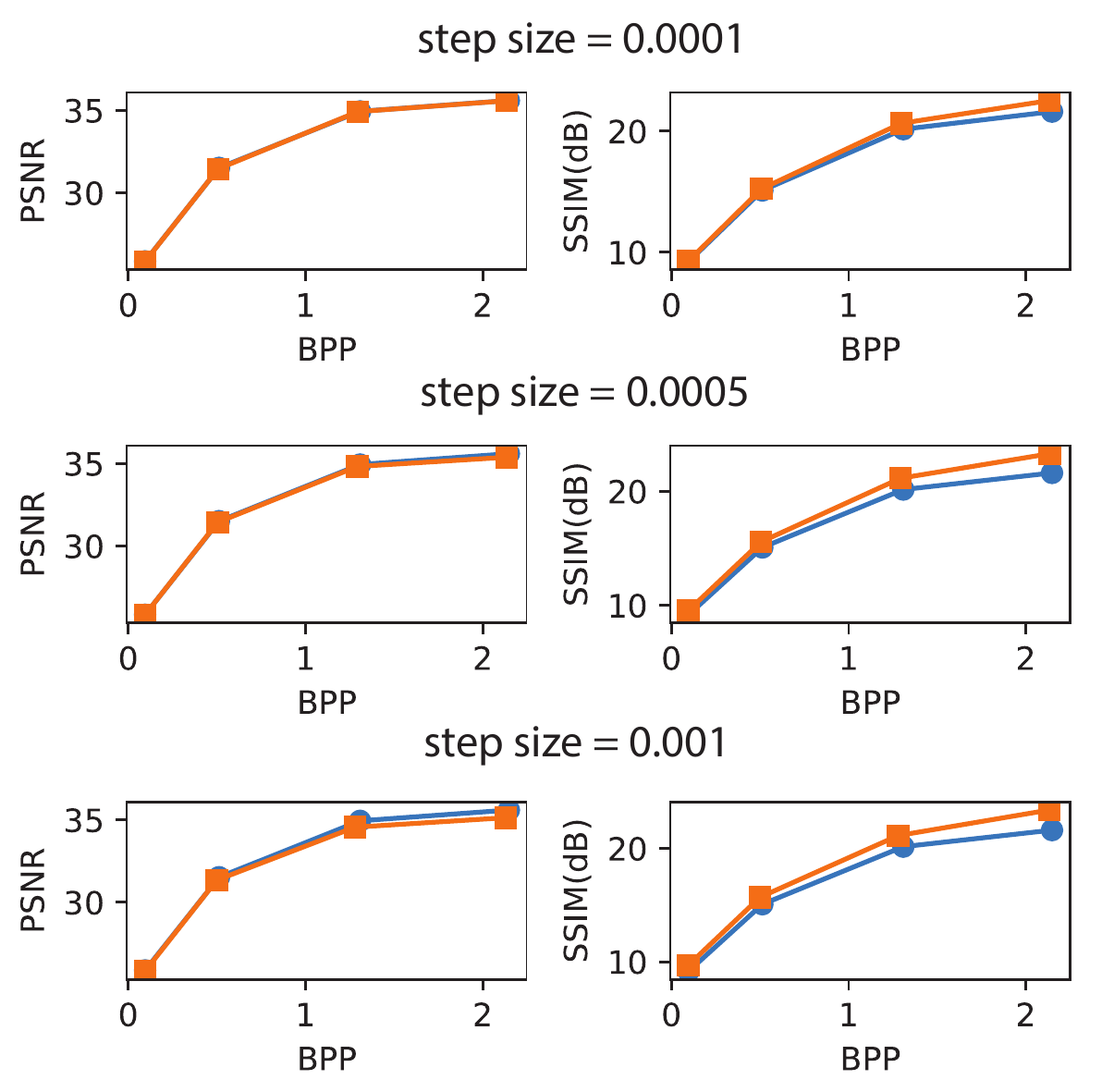}
    \caption{Effects on PSNR and MS-SSIM when increasing step size for generating substitutional images that target MS-SSIM performance. }
    \label{fig:appendix_targetssim}
\end{figure}

However, we found this drop on PSNR can be avoided by using a smaller step size in substitutional image generation, at the cost of a less significant improvement on MS-SSIM (see Fig.~\ref{fig:appendix_targetssim}).

\end{document}